\documentclass[twoside]{article}
\PassOptionsToPackage{numbers, compress}{natbib}
\usepackage{booktabs}
\usepackage[T1]{fontenc}
\usepackage[utf8]{inputenc}
\usepackage{array}
\usepackage{verbatim}
\usepackage{multirow}
\usepackage{amsmath, amssymb, amsthm}
\usepackage{graphicx}
\usepackage{fullpage}
\usepackage{epstopdf, amsfonts}
\usepackage{algorithmic, algorithm}
\usepackage{bbold}
\usepackage{url}
\usepackage[authoryear]{natbib}
\usepackage{float}
\usepackage{hyperref}

 \usepackage[accepted]{aistats2020}

\newtheorem{theorem}{Theorem}[section]

\newtheorem{lemma}[theorem]{Lemma}

\providecommand{\tabularnewline}{\\}

\begin{document}

\twocolumn[

\aistatstitle{Scalable Feature Selection for (Multitask) Gradient Boosted Trees}

%\aistatsauthor{ Anonymous Author(s)}
\aistatsauthor{ Cuize Han \And Nikhil Rao \And  Daria Sorokina \And Karthik Subbian  }
\aistatsaddress{Amazon, Palo Alto, CA \quad \{cuize, nikhilsr, dariasor, ksubbian \} @ amazon.com }
%\aistatsaddress{ cuize@amazon.com \And nikhilsr@amazon.com \And dariasor@amazon.com \And ksubbian@amazon.com } 
]

\begin{abstract}
Gradient Boosted Decision Trees (GBDTs) are widely used for building ranking and relevance models in search and recommendation. Considerations such as latency and interpretability dictate the use of as few features as possible to train these models. Feature selection in GBDT models typically involves heuristically ranking the features by importance and selecting the top few, or by performing a full backward feature elimination routine. On-the-fly feature selection methods proposed previously scale suboptimally with the number of features, which can be daunting in high dimensional settings. We develop a scalable forward feature selection variant for GBDT, via a novel group testing procedure that works well in high dimensions, and enjoys favorable theoretical performance and computational guarantees. We show via extensive experiments on both public and proprietary datasets that the proposed method offers significant speedups in training time, while being as competitive as existing GBDT methods in terms of model performance metrics. We also extend the method to the multitask setting, allowing the practitioner to select common features across tasks, as well as selecting task-specific features. %Experiments show that the proposed method outperforms traditional approaches. 
\end{abstract}

% !TEX root = gtgbm_aistats.tex

\section{Introduction}
\label{sec:intro}
Gradient Boosting methods \cite{friedman2001greedy} are widely used in several ranking and classification tasks for web-scale data \cite{zheng2008general, li2008mcrank}. GBDTs allow for efficient training and inference for large datasets \cite{ke2017lightgbm, chen2016xgboost}. Efficient inference is of key importance for applications such as search, where real-time vending of results at web scale in response to a search query is vital. %Search engines typically have tens of milliseconds between the user submitting a query, and the system vending the results. 
%Typical time constraints in search engines between the user submitting the query and the vending of results is usually in the order of tens of milliseconds. 

A key consideration for the models is the number of features used. A large number of features selected in the model severely impacts latency. Selecting a small number of features also allows for better model fitting and helps yield explainable models. While it is generally accepted that fitting a parsimonious model to the data is useful, past work on learning such models for GBDTs have been few and far between. In \cite{ke2017lightgbm}, sparsity inducing penalties are used to reduce the number of trees; in \cite{chen2016xgboost}, a similar technique is applied to penalize the number of leaves in each tree. One common method for feature selection in gradient boosting involves fitting the model on \emph{all} the features, ranking the features in the order of importance \cite{ke2017lightgbm, chen2016xgboost} and selecting the top-s, where $s$ is a positive, predefined number of features that one can handle. This kind of post-hoc thresholding is suboptimal compared to learning a sparse set of features during training itself. A second, and more often used method is backward feature elimination \cite{mao2004orthogonal}: recursively fit a model on the (leftover) set of features, and eliminate the least important feature. The second method becomes cumbersome in the case of most real world applications, which have a small number of target features and a large number of potential features to choose from. %when the number of target features is small, and the number of potential features to select from is huge, as in the case of most real world applications. %The function fitting procedure itself scales linearly with the number of samples, which again is large in most cases. 

To alleviate this, \cite{xu2014gradient} proposed a forward feature selection method for gradient boosting, based on a sparsity-inducing penalty over the features. The resulting subroutine to select features is linear in the number of features. This is both wasteful and cumbersome in high dimensional settings where the number of features we want to use is significantly smaller than the total number of features available.
%At each node of each tree that's being fit, the method performs the usual check over all the features to find the best split point, and penalizes features that have not yet been selected. Each of these subroutines is linear in the number of features.  This is both wasteful and cumbersome in high dimensional settings where the number of features we want to use is significantly smaller than the total number of features available. %Moreover, the method does not immediately carry over to using ranking losses as it does not account for the increased variance in the data as the number of trees fit increases \cite{burges2010ranknet}. 
Moreover, the sparsity penalty in the algorithm does not explicitly account for the distribution of targets in the training data for each tree. The difference in variance across the trees %can be big, so 
means a sparsity penalty that works well for one tree might not work well for subsequent ones.

In this paper, we help address both of the above concerns. We first show how the forward feature selection method for GBDT needs to be modified to account for %this increased variance% 
different variances in the residuals being fit, which we refer to as A-GBM (Adaptive Gradient Boosting Machine), since it adapts to the residual variance while fitting successive trees. The main contribution of our work is the introduction of a scalable variant of A-GBM, called GT-GBM (Group Testing GBM) that uses a group testing procedure to significantly speed up the training procedure for GBDTs. For cases where we want to select $s$ out of $d$ features, we show that so long as the number of samples in a node $n$ to split is at least the order of  $ \left( \frac{d}{s} \right)^2 \log \log\left( \frac{d}{s} \right)$, GT-GBM selects the optimal feature to split on. GT-GBM also enjoys computational speedups so long as $n$ is $O \left( \exp \left( \frac{d/s}{\log(d/s)} \right) \right) $.  Thus, so long as the rather easy-to-satisfy
\[
\left(\frac{d}{s}\right)^{2}\log\log\left(\frac{d}{s}\right)\lesssim n\lesssim \exp\left(\frac{d}{s} \log^{-1}\left(\frac{d}{s}\right)  \right), 
\]
condition holds, GT-GBM is guaranteed to be fast as well as accurate. This covers a wide range of real world applications. For example in web search cases, the number of samples is in the millions, number of features is in the hundreds and a few tens of features need to be selected.

Another major contribution is the extension of GT-GBM to the multitask setting, where our novel penalization helps us tradeoff between selecting common features across tasks, as well as task specific features. %We show that there's a tradeoff to be made between selecting the same set of feature across tasks, and separate features for each task. 
By sharing some features across tasks and selecting a few task-specific features, we can achieve better performance than standard multitask learning. We experimentally show that GT-GBM matches other feature selection methods for GBDT in performance, while being significantly faster. Results on multitask learning show the power of the flexibility to select features provided by our method. GBDT based feature selection has been shown to outperform other baselines such as the L1-regularized linear models and random forests \cite{xu2014gradient}, so we omit these redundant comparisons to those methods in this paper. %Links to our code will be made public in the paper's final version. 
%All our code is available on Github, and we will provide links in the final version of the paper. 
%We perform extensive experiments and show that the methods we develop outperform baseline methods significantly in terms of speed, while matching them for performance (such as AUC).  Results on multitask datasets show that our method's flexibility allows us to obtain state of the art performance when compared to the standard multitask setting of learning the same model across tasks, as well as the other extreme of training a separate model for each task. All our code is available on Github, and we will provide links in the final version of the paper. 

\paragraph{Prior Work :}
%Feature selection methods in linear models have been widely studied in the statistics and machine learning community. 
The LARS \cite{efron2004least} and Lasso \cite{tibshirani1996regression} methods, along with variants \cite{needell2009cosamp, chen2001atomic, rao2015forward} allow for highly efficient training and inference on large datasets for linear models. In the nonlinear setting, kernel methods \cite{song2012feature} can be trained with methods similar to the above ones, but their computational and memory complexity typically grow super-linearly with the number of samples in the data. The method in \cite{xu2014gradient} (referred to as GBFS, stands for Gradient Boosted Feature Selection) is a form of forward feature selection in the GBDT setting, but the tree splitting routine(s) still takes linear time with respect to the number of features in the data. We show how to avoid this. We also make a modification to GBFS to make the method more robust to the variances in the residuals as we fit more trees into the model. 

Multitask learning (MTL) \cite{caruana1997multitask} aims to improve model performance across multiple ``tasks" by learning joint representations. Such methods are useful in cases where there is not enough data to train individual models, as in the case of neuroscience \cite{rao2013sparse} or where there are similarities across tasks \cite{chen2010graph, yang2009heterogeneous}. Work on MTL has focussed on linear models \cite{maurer2013sparse}, where novel sparsity-aware penalties have been proposed to share models,  and neural networks (\cite{collobert2008unified} for languages for example); the former being too restrictive in web search and recommendations domain, and the latter not lending itself well to real-time inference. 

%The rest of the paper is organized as follows. 
We formally set up the problem we intend to solve and introduce the GBFS procedure of \cite{xu2014gradient} in Section \ref{sec:setup}, and the variance adaptive variant of the same. In Section \ref{sec:mtl}, we introduce our multitask learning method for forward feature selection. In Section \ref{sec:gtgbm} we derive a scalable method for forward feature selection in GBDT, and provide theoretical performance guarantees. We conduct extensive experiments in Section \ref{sec:experiments}, and conclude the paper in Section \ref{sec:conc}. 

\section{Problem Setup and GBFS}
\label{sec:setup}
%Uppercase and lowercase letters indicate matrices and vectors respectively. %Let $(x_i, y_i)_{i=1}^n \sim P(\mathcal{X}, \mathcal{Y})$ be a dataset of $n$ samples, sampled $iid$ from the joint probability distribution $ P(\mathcal{X}, \mathcal{Y})$, with $x_i \in \mathbb{R}^d$ being a d-dimensional vector. 
 Let $(x_i, y_i)_{i=1}^m$ be a dataset of $m$ samples, with $x_i \in \mathcal{X} \subset \mathbb{R}^d$. %We make no assumptions about $\mathcal{X}$.
   $y_i$ is a $T-$ dimensional vector (for the multitask case).
   % with $T=1$ corresponding to the standard single-task case that we address first.%
   Our aim is to train a GBDT model $f_{gbdt}(\mathcal{X}) \rightarrow \mathcal{Y}$, by using a small subset of features of size $s \ll d$. We denote by $[d]$ the set $\{1, 2, \cdots, d\}$. $\mathbb{1}\{C\}$ is the indicator function, taking the value $1$ if condition $ C$ is satisfied, 0 otherwise. 

Given $\mu > 0$, \cite{xu2014gradient} proposed the GBFS method, that penalizes the selection of new features via an additive penalty. %TODO2: The key step is penalizing features that have previously been selected via an adaptation of the capped $\ell_1$ norm \cite{zhang2009multi}.
%\[ 
%\|x\|_{capped-l1} := \sum_{j = 1}^p \min(1, | x_j | c_0 ), 
%\]
%where $c_0$ is a constant that depends on the number of samples, dimension and the noise level in the data.
Let $h$ correspond to a tree, and $\Omega \subset [d]$ be the set of features used by the model, and $g_i$ be the residual. At iteration $k$, GBFS solves
\begin{align}
\label{gbfs}
h_{k} &=\arg\min_{h\in{\mathcal H}}\sum_{i=1}^{m}\left(g_{i}-h({x}_{i})\right)^2 + \\
\notag
& \mu \sum_{j=1}^{d} \mathbb{1} \left\{h\text{ uses feature } j \text{ and }j\notin\Omega\right \}, 
\end{align}

with $\mathcal{H}$ being the space of trees we are optimizing over. \eqref{gbfs} can be solved by modifying the CART algorithm, which builds the tree by choosing the split to minimize the square error loss $L$. At each node, one chooses the best split among features $j\in[d]$
and split points $s_{j}\in\{x_{ij},\,i\in[m]\}$ that minimizes
\[
L(j,s_{j})=SSE_{L}(j,s_{j})+SSE_{R}(j,s_{j})+\mu  \mathbb{1}_{j}
\]
We have used the shorthand $\mathbb{1}_j$ to denote the indicator function for the event that feature $j$ has not been previously used.
\begin{align*}
SSE_{L}(j,s_{j}) &=\sum_{i}(y_{i}-\bar{y}_{L})^{2} ~\ \mathbb{1}\left\{ x_{ij}<s_{j}\right\}  \text{ and } \\
SSE_{R}(j,s_{j}) &=\sum_{i}(y_{i}-\bar{y}_{R})^{2} ~\ \mathbb{1}\left\{ x_{ij}\geq s_{j}\right\} 
\end{align*}
%\[SSE_{L}(j,s_{j})=\sum_{i}(y_{i}-\bar{y}_{L})^{2} ~\ \mathbb{1}\left\{ x_{ij}<s_{j}\right\} 
%\]
%and 
%\[
%SSE_{R}(j,s_{j})=\sum_{i}(y_{i}-\bar{y}_{R})^{2} ~\ \mathbb{1}\left\{ x_{ij}\geq s_{j}\right\} 
%\]
are the sum of squared errors for left and right child if we split
at feature $j$ and split points $s_{j}$. $\bar{y}_{L}=\frac{\sum_{i}y_{i} \mathbb{1}\left\{ x_{ij}<s_{j}\right\} }{\sum_{i} \mathbb{1}\left\{ x_{ij}<s_{j}\right\} }$,
$\bar{y}_{R}=\frac{\sum_{i}y_{i} \mathbb{1}\left\{ x_{ij}\geq s_{j}\right\} }{\sum_{i} \mathbb{1}\left\{ x_{ij}\geq s_{j}\right\} }$
are the means in the corresponding node. 

\paragraph{Adaptive Gradient Boosted Feature Selection :}
\label{sec:gbfsa}
%We first address a potential problem with the original method in \cite{xu2014gradient}. 
When optimizing to choose $h_{k}$, the value of the objective function in the root of the tree being built may have high variance across trees. Consequently, a penalty parameter $\mu$ that worked well until iteration $k-1$ might not be good for iteration $k$. Picking a good penalty parameter $\mu$ in this case becomes challenging, since we are using the same parameter for feature selection across all boosting rounds. To alleviate this situation, we propose to scale the
loss function being used to fit each tree to account for the current tree root variance. That is, we modify
 $L$ to be 
\[
\tilde{L}(j,s_{j})=\frac{SSE_{L}(j,s_{j})+SSE_{R}(j,s_{j})}{SSE_{r}}+\mu  \mathbb{1}_{j}
\]
 where $SSE_{r}=\sum_{i}(y_{i}-\bar{y})^{2}$ , $y_{i}$ is the label
in the current tree root. We now only need to choose $\mu \in [0, 1]$ since the scaled split criterion $\frac{SSE_{L}(j,s_{j})+SSE_{R}(j,s_{j})}{SSE_{r}}$ is always $ \in [0, 1]$. More importantly, this variance scaling ensures that the behavior of $\mu$ remains stable across each fitting round, avoiding the alternative of potentially ``re-tuning" $\mu$ for each boosting round.  We refer to this method as A-GBM (the `A' referring to adaptive), since the method adapts to the variance on a per-tree bases. A-GBM training proceeds exactly like GBFS, except for the scaling part. We refer the interested reader to Appendix \ref{app:agbm} for the pseudocode.

\section{Multitask A-GBM with Feature Selection}
\label{sec:mtl}
The above modification that adapts to the data variance as we grow the model becomes more crucial in the multitask learning setting, where we now have $T$ different but related tasks.  Let $t \in [T]$ denote the task id, and let the data for task  $t$ be $({x}_{i}^{t},y_{i}^{t})$, $i \in [m_{t}]$
and the corresponding features be $f_{j}^{t},\,j\in[d]$. For ease of presentation, we assume that all tasks have the same number of features $d$. In the case where the tasks have different features, we can `zero-pad' the data and since there is no variance along these features, they will not be considered for selection in the GBDT model. 

As in standard MTL, we can form groups of features, where each group is a single feature grouped across tasks \cite{maurer2013sparse}. Then we have $d$ groups of features $G_{j}=\{f_{j}^{t},t=1,.,T\}, j=1,2,..,d$. Assuming the tasks are related, grouping the features in this manner helps us learn a joint set of features that are useful across all tasks. However, this constraint might be too restrictive: we would like to account for slight variations across tasks, and have the flexibility to select task-specific features as well. To this end, we propose to use a group sparse penalty (only penalize if the feature is from a previously unused group of features, see the formula below for details)  $+$ sparse penalty for MTL. 
Note that now, the function to be fit depends on the task as well as the feature, giving:
\begin{align}
\label{mtlimp}
\tilde{L}(j,s_{j}) &=\frac{SSE_{L}(j,s_{j})+SSE_{R}(j,s_{j})}{SSE_{r}} \\
\notag
&+\mu_{G}I\left\{ j\notin\Omega_{G}\right\} +\mu_t I\left\{ j\notin\Omega^{t}\right\}, 
\end{align}

Where $\Omega_G$ is the set of features that have been selected across all tasks, and $\Omega^t$ is the set of features selected for task $t, ~\ t \in [T]$. $\mu_G, \mu_t$ are respectively the common group sparsity parameter for all the tasks and the task specific sparsity parameter.  The pseudocode for this method is presented in Algorithm \ref{mtlgbm}.

\begin{algorithm}[!h]
  \caption{Pseudocode for Multitask A-GBM}
   \label{mtlgbm}
  \begin{algorithmic}[1]
  \REQUIRE  Data $\{ {x}_{i}^{t},y_{i}^{t} \},\,i\in [m_t],$ $t \in [T]$, 
shrinkage $\epsilon$, iterations $N$, tree growth parameter $\alpha$, group penalty parameter $0 \leq \mu_{G} < 1$, individual task penalty parameter $0 \leq \mu_t <1 $, also
$\mu_{G}+\mu_t<1$
  \FOR{$t = 1, 2, \ldots T$}
  \STATE Initialize prediction $H^{t}=0$, residues $g_{i}^{t}=y_{i}^{t}$
and selected feature set $\Omega^{t}=\emptyset, \quad \Omega_{G} = \emptyset$ 
  \ENDFOR
  \FOR{$k = 1, 2, \ldots N$} 
  \FOR{$t = 1, 2, \ldots T$}
  \STATE Fit a tree $h_{k}^{t}$ using $\alpha$, $\mu_t$, $\mu_G$, data $\{\{{x}_{i}^{t},g_{i}^{t}\},\,i \in [m_t]\}$ and loss function \eqref{mtlimp}
  \STATE  $H^{t} = H^{t}+\epsilon h_{k}^{t} $
  \STATE  $ g_{i}^{t}=y_{i}^{t}-H^{t}({x}_{i}^{t}) $
  \STATE  $ \Omega^{t}=\Omega^{t}\cup\left\{ j~\ | \text{ tree }h_{k}^{t}\text{ uses feature }f_{j}^{t}\right \} $
  \STATE  $ \Omega_{G}=\Omega_{G}\cup\left\{ j ~\ | \text{ tree }h_{k}^{t}\text{ uses feature }f_{j}^{t}\right \} $
  \ENDFOR  
  \ENDFOR
     \STATE Output  $H^{t}$, $\Omega^{t}  ~\ \forall ~\ t \in [T]$ and $\Omega_{G}$
   \end{algorithmic}
\end{algorithm}

Using a combination of the group sparse and sparse penalizations has been shown to be effective in multitask learning settings for linear regression and classification \cite{rao2013sparse, simon2013sparse}. To the best of our knowledge, this has not been proposed before in the tree learning setting. 

\section{Scalable adaptive Gradient Boosting}
\label{sec:gtgbm}

The methods described above end up having to compute the $SSE_{L}(j,s_{j})$ and $SSE_{R}(j,s_{j})$ functions defined previously for all the features in the dataset and for each split to be performed while fitting a tree. This procedure is linear in the number of features $d$, and the number of samples $n$ per node where the split is being computed. For many real world applications in web search and recommendations, the total number of feature is large while the number of feature used by the model is significantly smaller. In these cases, we expect that checking all the $d$ features is not only time consuming but also redundant. If we can quickly identify those small number of $s$ good features \emph{without} checking them all during each node
split, training time will be reduced greatly. We address this now.  

\subsection{Group Testing and Binary Search}

The idea is to compare groups of randomly selected features and perform a binary search to eliminate the set of features that are relatively uninformative. Random selection helps reduce the bias in the ordering of the features. At each time
we can eliminate half of features in this way. This depends on a key consideration: we require a metric that can be computed \emph{efficiently} on a group of features, and one that is also \emph{indicative} of the presence of an important feature in the group. An inefficient method will not yield computational gains, and a non-indicative metric is not going to yield an accurate solution. Suppose
we have a function \texttt{GT($G$,$M$)} that takes in a subset
of features $G \subset [d]$ and a subset of samples $M \subset [m]$ as input arguments. Suppose the number of operations it takes to evaluate a split for this group of features is $\Phi(|G|,|M|)$: the computational complexity of this procedure depends on the number of samples as well as the number of features.

We will address how to construct such a function in Section \ref{sec:gtfunc}. Assuming for now we do have such a function at our disposal, we give our general procedure of group testing and binary
search in GBDT. We refer to this method as GT-GBM, the ``GT" referring to the group-testing scheme. The pseudocode for GT-GBM is identical to that of A-GBM (Algorithm \ref{agbm}) except line \ref{minfunc} will be replaced by the subroutine we provide below in Algorithm \ref{ffs}. For the multitask case, line \ref{mtlimp} in Algorithm \ref{mtlgbm} will be replaced by the subroutine.

\begin{algorithm}[!h]
  \caption{Tree Fitting Subroutine for GT-GBM}
   \label{ffs}
%  %Regularization parameter:\lambda, Step size $\eta$}
  \begin{algorithmic}[1]
  \REQUIRE (in addition to usual hyperparameters) $s$ = desired number of features,  $\delta\in (0,1)$  (see Theorem \ref{thm:speed} for details)
   % \STATE Initialize $\tilde{s} = \max(s - \hat{s}, 1)$\footnote{the algorithm always try to select new features, the total number of feature selected should depend on $\mu$ not $s$, The latter is only used to determine when will the algorithm triggers the group testing and binary search procedure }
%    \IF{$\tilde{s} < \tau$}
   	 \STATE Check previously used feature set $\Omega$  for splitting and record the best standardized MSE. Call it $l$.
   	\STATE Independently generate $es\log(\frac{s}{\delta})$ random subsets
from $[d]$ with size $\frac{d}{s}$ . If $s=1,$we just select $[d]$ . Assign these to $\mathcal{G}$
	\STATE Initialize candidate set $C = \emptyset$
	\FOR{Each random subset $G \in \mathcal{G}$}
	\WHILE{$|G| > 1$}
		\STATE Binary half split G into $G_L, G_R$. Let $n_G$ be the samples to consider for this split.
		\STATE $G = \arg \min_{G_L, G_R} \left( \texttt{GT($G_{L}$,$n_G$)} ,  \texttt{GT($G_{R}$,$n_G$)} \right) $
 	\ENDWHILE
 	\STATE $C = C \cup G$
 	\ENDFOR 
 	\STATE check features in $C$ for splitting and record the squared error value with penalty
$l'+\mu$ if the feature is not used by previous trees. 
	\IF{$l' + \mu < l$}
	\STATE Include best feature from $C$
	\ELSE
	\STATE Use one of the old features from $\Omega$ to split
	\ENDIF
 %\ELSE
 %   \STATE fit a tree exactly as in Algorithm \ref{agbm}, step \ref{minfunc}
%    \ENDIF
    %\STATE Update $H, g, \Omega$ similar to Algorithm \ref{agbm}
  %\STATE Output  $H$ and $\Omega$
   \end{algorithmic}
\end{algorithm}

\subsection{Constructing an Efficient \texttt{GT()} Function}
\label{sec:gtfunc}

In general, the number of operations for group testing and binary
search in a node splitting step is $O\left(s\log(s)\log(\frac{d}{s}) \Phi(d,n)\right)$. Our aim is to construct a function \texttt{GT()} such that $\Phi(d,n) \ll O(nd).$ %If the complexity of the \texttt{GroupTest} function $\Phi(d,n)$ is small, then this value will be smaller than $O(nd)$. 
We do this as follows: given a group of features and the samples to make the split, we sum the features up to obtain a new ``pseudo-feature'' \footnote{We will standardize feature values by subtracting the min value and dividing the max value, so that all feature values are within [0,1]}. We will then test this ``pseudo-feature'' for a split point in a fashion identical to the usual tree-splitting procedure in A-GBM. For speeding up this computation, we can compute the prefixed-sum  \cite{cormen2009introduction} of all the features in data \footnote{after computing the prefixed sum for features in each random subset and storing the result in one-pass, getting the "pseudofeature" value will just be O(1)}.  $\Phi(d,n)$ now is $n\log(n)$ for sorting
the pseudofeature and the check for splitting. Comparing GT-GBM with the usual procedure of GBDT, we will gain
a boost in training speed if (see table \ref{tab:computation_complexity} for details) 
\begin{equation}
\label{nlb}
s\log(s)\log(n) << \frac{d}{\log(\frac{d}{s})},
\end{equation}
which is easy to satisfy in real world applications. More detailed complexity comparision is the following.
%a condition which is rather easily satisfied in most real world applications with large number of features $d$. %For example, with $d = 500, s = 20$ we need $n \leq 10^{7}$. Even with $d = 10000$ and $s = 100$, we need $n \leq 10^{12}$ for GT-GBM to be faster than A-GBM, and by extension other GBDT methods. 
 
The main preprocessing of data for training A-GBM or any other sort-based GBDT algorithm involves getting and storing the sorted value pairs of features and targets. This takes $O\left(\left(n\log n\right)d\right)$ operations and needs $O\left(nd\right)$ space. GT-GBM, however, does not need to precompute the sorted value pairs since target values will be sorted based on the pseudofeature during binary search and split. Instead, it calculates and records the prefixed sum for each of $O\left(s\log s\right)$ random subset of features with size $d/s$. So the precomputation for GT-GBM takes $O\left(\left(s\log s\right)nd/s\right)=O\left(\left(n\log s \right)d\right)$ time and space. With a bit more space used during precompute, GT-GBM needs $O\left(n\right)$ instead of $O\left(nd\right)$ space since sorted value pairs doesn't need to be  stored and passed to child nodes during growing the tree.  Table \ref{tab:computation_complexity} shows this comparison.

\begin{table}
\tiny
\caption{\label{tab:computation_complexity}Complexity comparisons between A-GBM (and hence GBFS) and GT-GBM. P and T refer to the precomputation and training phases respectively. }
\begin{tabular}{|c|c|c|c|}
\hline 
\textbf{Algo.}  & \textbf{Phase}  & \textbf{Time}  & \textbf{Space} \tabularnewline
\hline 
\multirow{2}{*}{A-GBM}  & P  & $O\left(\left(n\log n\right)d\right)$ & $O\left(nd\right)$ \tabularnewline 
 & T  & $O\left(nd\right)$  & $O\left(nd\right)$ \tabularnewline
\hline 
\multirow{2}{*}{GT-GBM}  & P  & $O\left(\left(n\log s \right)d\right)$ & \textbf{$O\left(n d\log s\right)$} \tabularnewline 
 & T &  $O\left(\left(s\log(s)\log(d/s)\right)n\log n\right)$ & $O\left(n\right)$ \tabularnewline
\hline 
\end{tabular}
\end{table}

\subsection{Theoretical Guarantees for GT-GBM}
If $s = 1$, then the important feature will be in either $G_L$ or $G_R$. All other features will act as random noise. Intuitively the procedure will select the group that contains the relevant feature with high probability as long as it is highly correlated with the target. This process recurses until we find the important feature.
%This process repeats until we find the right feature, noting that at each split point, the single right feature will be in $G_L$ or $G_R$.  
If there are multiple relevant features in the same group, however, their effects can cancel each other out.  An idea then is to generate several random subsets of features and apply our \texttt{GroupTest} to these subsets, as we do in Algorithm \ref{ffs} (line 2). If a subset contains only one of the important features, then it reduces to the case for one feature and we can find that feature with high probability.  The following result bounds this probability as a function of the number of subsets generated:

\begin{theorem}
\label{thm:speed}
Suppose that there are $s$ important features. To ensure that for every important feature there is a random subset that only cover this feature with probability $1 - \delta$, it is sufficient to generate $p$ random subsets of features, where
$
p \geq e s \log \left( \frac{s}{\delta} \right)
$
and $e=2.71..$  is the base of the natural logarithm. 
\end{theorem}
We refer the reader to Appendix \ref{app:proofspeed} for the Proof. %The bound in Theorem \ref{thm:speed} is quite conservative shown by the following experiment on synthetic data (see Appendix \ref{sec:ExpOn1Split} for details). We fix $n, d = 5000, 200$ and vary $s, \delta$. We randomly generate $p = es \log(s/\delta)$ subsets from $[d]$, apply group testing and binary search on each random subset and get $s \log(s/ \delta)$ candidate features. We record for each $s, \delta$ pair the fraction of total correctly selected features. Figure \ref{ptplot} shows the result of this procedure, averaged 50 times. We see that $s \sim 10 \delta$ can be used as a good rule to select the number of candidate features, thereby indicating $p = O(s)$ groups generated for GT-GBM is good enough. 

Next, we show that the method we proposed is guaranteed to recover the correct set of features with high probability, under mild asumptions.

%\begin{figure}
%\centering
%\includegraphics[width = 55mm, height = 35mm]{PT1.png}
%\caption{Fraction of the true features selected as a function of number of true features $s$ and input success probability $1-\delta$. Dark regions indicate values near 0, and light closer to 1. We see that we can choose $s/\delta = O(1)$ %to get reasonable performance, indicating $p \sim O(s)$ is a good choice to pick. The leftmost column $(s=1)$ shows that the binary search procedure basically can always find the single right feature.}
%\label{ptplot}
%\end{figure}

\begin{theorem} \label{thm:correct}
Suppose $X = (X_1, \ldots, X_d)$ are independent of each other, and $0 \leq X_i \leq 1$, with non-zero variance. Let $B_d^{2} = \text{Var}(X_1+..+X_d)$.  %and $\mu_d = \mathbb{E}(X_1+..+X_d)$%.
Assume $\lim_{d\rightarrow\infty}{B_d}=\infty$. %and $\eta=\lim_{d\rightarrow\infty}{\frac{\mu_d}{B_d}}$%. 
Suppose there is an unknown subset $S^{*}\subset[d], ~\ |S^{*}|=s$, such that 
$%\begin{equation}
Y=\mu+\sum_{i\in S^{*}}f_{i}(X_{i})+\epsilon,   ~\  \label{eq:additive form 1}
$%\end{equation}
where $\mu=\mathbb{E}Y$ is the population mean and $\epsilon$ is noise (mean 0, bounded and independent of all other variables). $f_{i}$s are unknown univariate monotonic functions with $\mathbb{E}f_i(X_i)=0$. Suppose at a node we have $n$ i.i.d. samples. Then for $\delta \in(0,1)$, if $d\geq d_{0}$ and 
\begin{equation}
n \geq C_{0}\left(\frac{d}{s}\right)^{2}\log\left(\log\left(\frac{d}{s}\right)\frac{\log(1/\delta)}{\delta}\right)\label{eq: theorem condition 1}
\end{equation}
GTGBM finds the best split feature with probability at least $1-\delta.$, where $C_{0}$ and $d_{0}$ are positive constants that only depend on the fixed unknown functions $f_{i}$, $i\in S^{*}$. 
%and $\eta$%.
\end{theorem}
Note that the assumptions made above are based on Sparse Additive Models \cite{ravikumar2009sparse}, and encompass a wide variety of practical settings. 

\begin{proof}[{\bfseries Proof Sketch}]
	Recall the split criterion of CART algorithm. For a split variable $Z$ (a feature or the pseduo-feature in GTGBM that represents a group of variables) and threshold $t$, the criterion is to minimize 
	\begin{equation}
	L_{n}(Z,t)=\frac{1}{n}\left(\sum_{i:Z_{i}<t}\left(Y_{i}-\bar{Y}_{L}\right)^{2}+\sum_{i:Z_{i}\geq t}\left(Y_{i}-\bar{Y}_{R}\right)^{2}\right)\label{eq: empirical split 1}
	\end{equation}
	The population split criterion (corresponds to when we have infinite amount of data) is to minimize
	\begin{eqnarray}
	L(Z,t) & = & \mathbb{E}[\left(Y-\mathbb{E}\left[Y|Z<t\right]\right)^{2}1\{Z<t\}\nonumber \\
	&&+\left(Y-\mathbb{E}\left[Y|Z\geq t\right]\right)^{2}1\{Z\geq t\}]\label{eq: theoretical split 1}
	\end{eqnarray}
	For an important feature index $i \in{S^*}$, we consider the random subset $S$ generated in GTGBM that only covers $i$. Then during binary search for active feature within $S$,  we only need to prove for the split subset $S_L,S_R$ (assume $S_L$ contains the important index $i$), that $\min_{t}L_{n}(Z_{S_L},t)<\min_{t}L_{n}(Z_{S_R},t)$ w.h.p. %with high probability. 
	
	Let $Z_S=\sum_{j\in S}X_j$ . For the population version, we can prove $L(Z_{S_R},t)=\mathbb{E}Y^2,\forall t$ (no variance reduction),  $\min_{t}L(Z_{S_L},t)<\mathbb{E}Y^2$ and the difference only depends on the signal strength of $f_i$ and how correlated are $Z_{S_L}$ and $Y$.  To investigate the sample split criterion, we need to quantify : (a) How the amount of variance reduced decays with the increase of  $| S_L |$ %$S_L$'s size (since $Z_{S_L}$ will be less correlated with $Y$)? 
	(Lemma \ref{app: therotical split lemma} states %it's the inverse of it's size:  
	$\approx\frac{1}{\left|S_{L}\right|}\approx\frac{s}{d}.\label{eq: signal decay 1}$) 
		(b) How the uniform approximation error between empirical and population split criterion decays with $n$. (Lemma \ref{app: emprical split lemma} states $\sup_{t}\left|L_{n}(Z,t)-L(Z,t)\right|=O_p(\sqrt{\frac{ 1}{n}}) $)

Combining the above gets us the result. We refer the reader to Appendix \ref{app:proofcorrect} for the detailed proof. 
\end{proof}

%we see the condition for GTGBM to successfully find the important feature is $\frac{s}{d} \gtrsim\sqrt{\frac{\log n}{n}}$ which yields $n\gtrsim\left(\frac{d}{s}\right)^{2}\log\left(\frac{d}{s}\right)$ as stated in the theorem.  We refer the reader to Appendix \ref{app:proofcorrect} for the full detailed proof. 
Combining equations \eqref{nlb} and \eqref{eq: theorem condition 1} in Theorem \ref{thm:correct} show that so long as the number of samples $n$ at a node to split satisfies
\[
\left(\frac{d}{s}\right)^{2}\log\log\left(\frac{d}{s}\right)\lesssim n\lesssim \exp\left(\frac{d}{s} \log^{-1}\left(\frac{d}{s}\right)  \right), 
\]
GT-GBM will find the correct feature to split significantly faster than GBFS. This condition is easily satisfied in most real world applications, where the number of samples and the number of features are large, and relatively shallow trees are used to train the models which is the case for gradient boosting procedures. %Next, we empirically verify that GT-GBM indeed outperforms GBDT baselines. 

\begin{figure}[H]
\centering
\includegraphics[width = 70mm, height = 50mm]{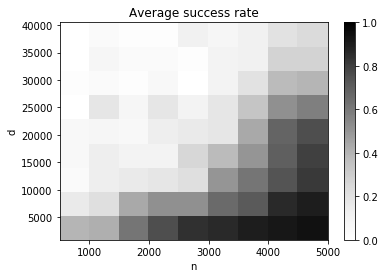}
\caption{Average success rate as a function of ambient dimension $d$ and sample size $n$. Dark regions indicate values near 1, and light closer to 0. Note the near linear dependence between $n$ and $d$.} 
\label{ptplot}
\end{figure}

An experiment on synthetic data shows the bound in Theorem \ref{thm:correct} is quite conservative. Figure \ref{ptplot} indicates that the dependence between $n$ and $d$ is potentially linear. We leave the tightening of the bound for future work. For the experiment, we fix $s=3,\delta=0.1$ and generate $y=2x_1-3*2^{x_2}+\log_{2}(1+x_3)+\epsilon$ where $x_1, x_2, x_3$ and other irrelevant features are i.i.d uniform on $[0,1]$ and $\epsilon \sim \mathcal{N}(0, 1)$. We replicate each experiment 50 times and calculate the ratio of success (success means the candidate feature set found by GTGBM contains both $x_1,x_2,x_3$). 

\section{Experiments and Results}
\label{sec:experiments}
First, we extensively test A-GBM and GT-GBM on publicly available datasets. Next, we apply the methods to proprietary datasets, and evaluate GT-GBM for ranking and multiclass classification tasks. Results on an internal dataset for classification are provided in Appendix \ref{app:internal_clf}. We compare our methods with other GBDT feature selection methods, as that is the main focus in this paper. %\cite{xu2014gradient} showed that these methods outperform other feature selection methods based on Random Forests and L1 regularized linear methods, so we omit these comparisons. 

\subsection{Public Datasets and Baselines}
We compare A-GBM and GT-GBM methods with GBFS \cite{xu2014gradient} and the GBDT method with ranking all features, and retraining with $K$ most important features (referred to as GBDT-topK here). For GBDT-topK, we use LightGBM  \cite{ke2017lightgbm} and use it's default feature scoring mechanism to rank the features by importance. We train the models on the Gisette\footnote{\url{https://archive.ics.uci.edu/ml/datasets/Gisette}} , Epsilon\footnote{\url{https://www.csie.ntu.edu.tw/~cjlin/libsvmtools/datasets/binary.html}}, and the Flight Delay\footnote{\url{https://github.com/szilard/benchm-ml}} datasets. They are all for classification tasks. For the latter, we use the variant with 100K samples, and the same script to generate the data as provided in the repository. 
% and the Yahoo Learning to Rank (YLTR) data \cite{chapelle2011yahoo}. 
%and the latter is a learning to rank dataset. 
Details for all the datasets are provided in Table \ref{tab:datasets}. 
%For reproducing the results, we provide the optimum hyperparameter values for all the methods in Appendix \ref{app:exptresults}.

\begin{comment}
\begin{table}[!h]
\
\begin{tabular}{|c|c|c|c|c|}
\hline 
 \textbf{Dataset} & \textbf{$\#$ samples}  & \textbf{$\#$ features}  & \textbf{Task} \tabularnewline
\hline 
\hline 
 \textbf{Gisette}  & 6000  & 5000  & Classification  \tabularnewline
\hline 
 \textbf{Epsilon}  & 80000  & 2000  & Classification  \tabularnewline
\hline 
 \textbf{Flight}  & 100000 & 634  & Classification \tabularnewline
\hline
\end{tabular}
\end{table}
\end{comment}

\begin{table}[!h]
	\caption{\label{tab:datasets} Experimental datasets }
	\begin{tabular}{|c|c|c|c|}
		\hline 
		\textbf{Dataset} & \textbf{$\#$ samples}  & \textbf{$\#$ features}  \tabularnewline
		\hline 
		\hline 
		\textbf{Gisette}  & 6000  & 5000    \tabularnewline
		\hline 
		\textbf{Epsilon}  & 80000  & 2000   \tabularnewline
		\hline 
		\textbf{Flight}  & 100000 & 634  \tabularnewline
		\hline
	\end{tabular}
\end{table}

For each of the methods we use, we tune all the parameters on a held out validation set, and report the results on a separate test set. For GBFS, A-GBM and GT-GBM, we choose the corresponding $\mu$ that achieves the best performance on the validation dataset, regardless of the number of features they select. For this reason, we end up picking different number of features for different methods. For GBDT-topK, we train on all the features, and pick top K features, where K is the maximum of the number of features picked by the 3 other methods. We then retrain the model with these K and report results on the test set. Optimal hyperparameter values to reproduce our results are provided in Appendix \ref{app:exptresults}. 
\vspace{-3mm}
\paragraph{Speed and Performance Comparisons :}
\label{compbaseline}
First, we show that the proposed methods perform either comparatively, or outperform the baselines. Table \ref{tab:results} shows the performance metrics for the methods we compare, indicating that there's very little performance loss over the baseline methods. For the sake of completeness, we also report the results obtained from training the GBDT model on all the features, with no feature selection in Appendix \ref{app:allfeatures}. Furthermore, the flight delay dataset has a large number of categorical features, and a large number of data points compared to features. Even in this case, GT-GBM outperforms the other baselines. 

\begin{table}[H]
\tiny
\caption{ \label{tab:results} Performance comparison on various datasets. Note that GT-GBM consistently picks fewer features while still outperforming or competing with A-GBM and GBFS. As expected, GBDT-topK suffers from poor approximation as a result of picking top K features after fitting on the whole set of features.}
\begin{tabular}{|c|l|c|c|c|}
\hline
\textbf{Dataset}	& \textbf{Method}	& \textbf{$\#$ feats}	& \textbf{RMSE}	& \textbf{AUC\_ROC} \\
\hline
\hline
			& GBDT-topK		& 178				& 0.187			& 97.88 \\
\textbf{Gisette}	& GBFS			& 172				& 0.183			& 99.01 $(+1.15 \%)$ \\
			& A-GBM			& 178				& 0.182			& 99.18 $(+1.33 \%)$\\
			& GT-GBM		& 170				& 0.182			& 99.19 $(+1.34 \%)$\\
\hline
			& GBDT-topK		& 306				& 0.377			& 91.8 \\
\textbf{Epsilon}	& GBFS			& 306				& 0.363			& 93.0 $(+1.99 \%)$\\
			& A-GBM			& 250				& 0.366			& 93.2 $(+ 2.21 \%)$\\
			& GT-GBM		& 255				& 0.373			& 93.2 $(+2.21 \%)$\\
\hline
			& GBDT-topK		& 67					& 0.391			& 71.1 \\
\textbf{Flight}	& GBFS			& 67					& 0.389			& 71.6 $(+0.70 \%)$\\
			& A-GBM			& 48					& 0.389			& 71.7 $(+0.84 \%)$\\
			& GT-GBM		& 45					& 0.390			& 71.6 $(+0.70 \%)$\\
\hline
\end{tabular}
\end{table}

\begin{figure*}
\centering
\includegraphics[width = 50mm, height = 30mm]{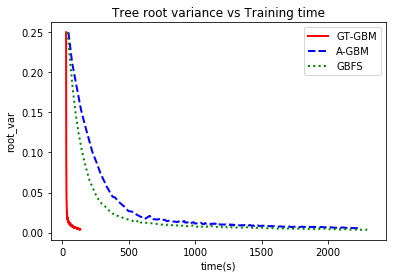}\includegraphics[width = 50mm, height = 30mm]{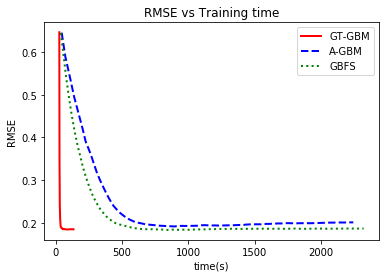}\includegraphics[width = 50mm, height = 30mm]{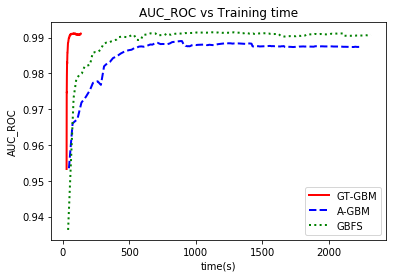}
\includegraphics[width = 50mm, height = 30mm]{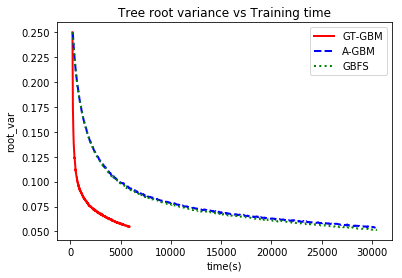}\includegraphics[width = 50mm, height = 30mm]{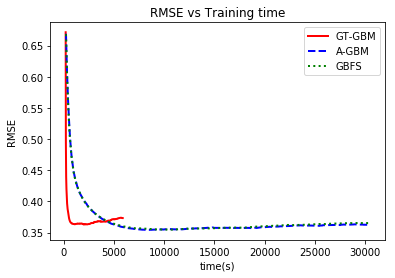}\includegraphics[width = 50mm, height = 30mm]{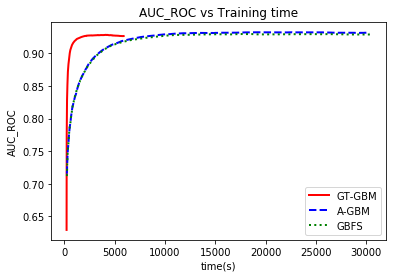}
\includegraphics[width = 50mm, height = 30mm]{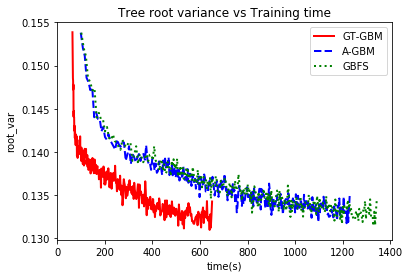}\includegraphics[width = 50mm, height = 30mm]{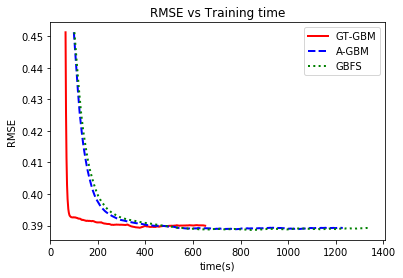}\includegraphics[width = 50mm, height = 30mm]{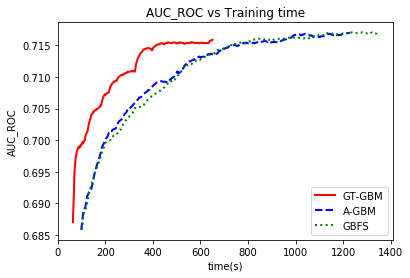}
\caption{ Timing comparisons of all the methods on various datasets, Gisette (top), Epsilon (middle), and Flight (bottom). In all the cases, we see that GT-GBM outperforms the other methods, by orders of magnitude. We plot the tree root variance (left), RMSE (middle) and Area under ROC curve (right) for all datasets as a function of time.}
\label{fig:time}
\end{figure*}

Next, we compare the training time for all the methods in Figure \ref{fig:time}. %for the classification datasets we consider.
%and the same for the learning to rank data in Figure \ref{fig:timeltr}. 
The Figure shows that GT-GBM is significantly faster than the competing methods on all the datasets, by an order of magnitude for Gisette, and two orders of magnitude for Epsilon. The gap is smaller for Flight dataset, since the ratio of the number of samples to the number of features is much smaller.

\vspace{-3mm}
\paragraph{Evaluating Correlations :}
%In this subsection, we empirically show that the GBDT method leads to suboptimal solutions when compared to GT-GBM, or a method that explicitly looks for the top features to select. We use the Gisette dataset for all our results here. 

\begin{figure}[H]
\centering
\includegraphics[width = 35mm, height = 30mm]{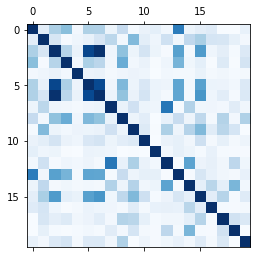}\includegraphics[width = 35mm, height = 30mm]{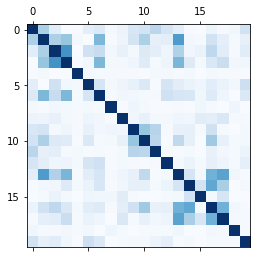}
\caption{Pairwise pearson correlations for the top 20 features selected by GBDT-topK (left) and GT-GBM (right) methods. The lighter squares indicate values closer to 0.} %GT-GBM selects more uncorrelated features, obtaining a more informative model with the same number of selected features. }
\label{fig:pcgisette}
\end{figure}

In Figure \ref{fig:pcgisette} we show that the features selected by the GT-GBM methods are less correlated than those picked by fitting all the features, and selecting the top K (via the feature importance scores obtained via GBDT). We fix $K=20$, and plot the Pearson correlation coefficient for the Gisette data. When the number of features we want to select is constrained, it is important to select features that are as uncorrelated from each other as possible, as this allows for maximal information gain. 

\subsection{Performance on Proprietary datasets}
Next, we apply the GT-GBM and A-GBM methods on proprietary datasets.  We use aggregated data sets containing only de-identified data from search logs of an e-commerce engine (i.e. they don't include personally identifying information about individuals in the dataset).  We make use of 4 datasets across 2 tasks. C1 and C2 are classification tasks, and R1 and R2 are ranking tasks. Results on C1 and C2 are in Appendix \ref{app:internal_clf}, since the previous experiments already evaluated GT-GBM on classification data. In all the cases below, we choose 20 as the desired number of features in our models so as to illustrate an example where extreme latency constraints are enforced.  \begin{comment}In all the cases below, the desired number of features in our models is $\sim 20$, due to extreme latency constraints in vending the models in real time. (ravi commented as inappropriate)  \end{comment} All the hyperparameters  for all the methods compared were tuned via cross validation, and we report the results on a held out test set. Since from Section \ref{compbaseline}, we see that the performance of A- and GT-GBM methods are similar, we only report the latter, and compare it to GBDT-topK and GBFS. %We optimize the RMS error in all the methods. 

The ranking task is akin to the standard relevance task in a search engine: in response to a query, and a set of items that are matched, the job is to rank the items in the order of relevance. Since this is a ranking task, we report the Mean Reciprocal Rank (MRR) for the datasets. Again, we see that GT-GBM is competitive with the other methods (while being faster) (Table \ref{tab:r}). 

\begin{table}[!h]
\tiny
\caption{\label{tab:r}Comparison of various methods on the Ranking tasks (R1 and R2). Similar to the classification setting, GT-GBM is competitive with the baselines, and achieves the same result in significantly less time.}
\begin{tabular}{|c|c|c|c|c|c|}
\hline 
\textbf{Dataset} & \textbf{Measure} & \textbf{GBDT-topK}  & \textbf{GBFS}  & \textbf{GT-GBM} \tabularnewline
\hline
\hline 
\textbf{R1} & \textbf{MRR}  & 0.530  & 0531  & \textbf{0.532} \tabularnewline
\hline 
& \textbf{RMSE}  & 0.159  & \textbf{0.158}  & \textbf{0.158} \tabularnewline
 \hline
 \hline 
\textbf{R2} & \textbf{MRR}  & 0.496  & \textbf{0.499}  & 0.498 \tabularnewline
\hline 
& \textbf{RMSE}  & 0.103 & \textbf{0.101}  & \textbf{0.101} \tabularnewline
\hline 

\end{tabular}
\end{table}

\subsection{Multitask Feature Selection}

Finally, we test the multitask variant of our algorithm on two other %Amazon 
proprietary datasets: M1 and M2. M1 is a classification dataset that categorizes a query into 3 categories (head, torso, tail). The idea is to see if there are highly predictive features in one task that can be used in other tasks where there is a lack of data. At the same time, there might be task-specific features that are useful, which our model accounts for as well. M2 is a dataset that uses query-items across countries, similar to the dataset used in \cite{chapelle2010multi}. Due to space constraints, details about M2 and results are provided in Appendix \ref{app:m2}.  

We tune the two parameters $\mu_{G}$ and $\mu_t$  which control the proportion of common active features and task-specific important features via cross-validation, and report the results on a held out test set. In Figure \ref{mtlm1} (and  \ref{mtlm2} in Appendix), SingleTask refers to training the model on the combined training data in the single task mode with the task number used as a categorical feature. Multitask\_GroupSparse refers to the Multitask model we developed, but forcing all the features across tasks to be the same, which is the standard multitask learning framework (effectively $\mu_t = 0$). Multitask refers to the model that has the full flexibility, where both sparse and group sparse parameters can be nonzero. ``Total" refers to the overall metric, after taking a weighted average of the scores across the tasks, weighted proportional to the number of samples in each task. The figures show that the Multitask model outperforms both the other methods, across all tasks as well as overall.  %Each task is a classification task.
%The weights are proportional to the number of samples in each task. Note that the ``Total" performance is skewed by the number of samples for each task, which might not be the best means to average since we will weight those tasks with more samples. In fact, the ``Single Task" algorithm will train a somewhat similar model, which we are looking to avoid here. 

\begin{figure}[H]
\centering
\includegraphics[width = 70mm, height = 40mm]{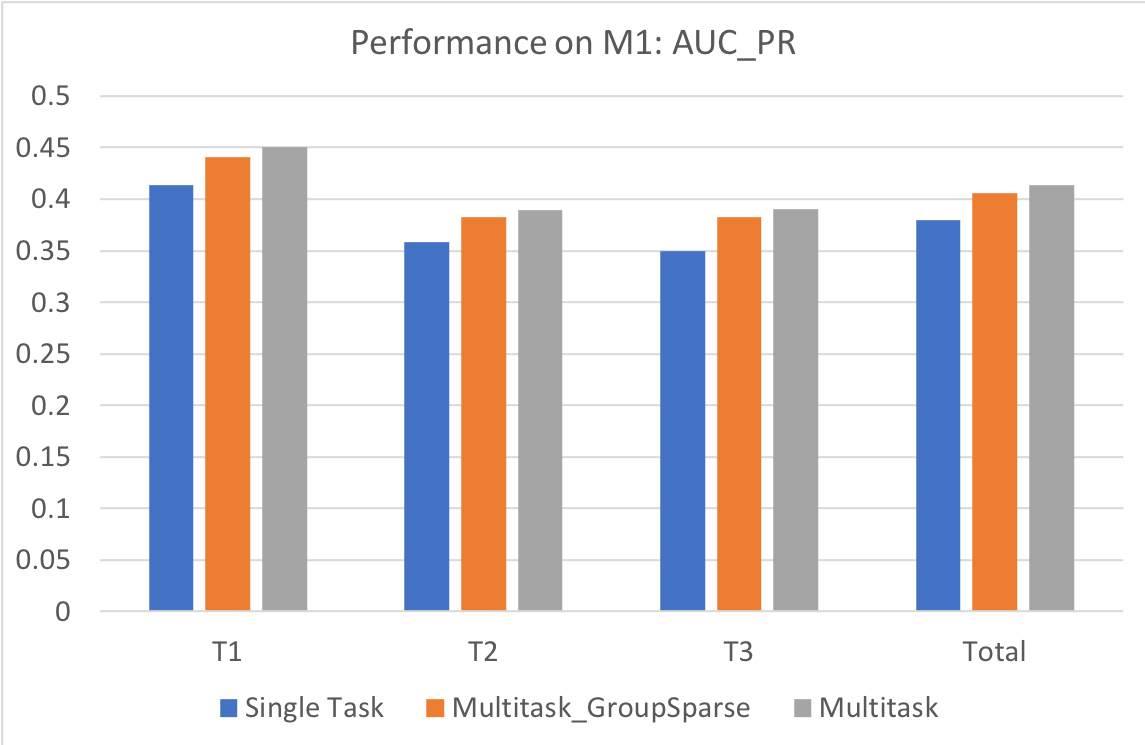} 
\caption{Performance on M1, for Area Under Precision-Recall curves. We see that having the flexibility to choose both task specific and common features across tasks helps boost performance. T1, T2, T3 refer to the three query level tasks respectively. }
\label{mtlm1}
\end{figure}

\section{Conclusions}
\label{sec:conc}
In this paper, we developed a feature selection procedure for gradient boosted decision trees that adapts itself to the variations in the data, and built a scalable version of the same. The scalable algorithm we developed uses a novel group testing and binary search heuristic to achieve significant speedups over baseline methods, with almost no change in performance. We provided theoretical performance guarantees that establish both the speedup and correctness, and empirical results corroborating the same. We also developed a multitask variant of this algorithm, that is flexible enough for the practitioner to transition between choosing the same set of features and training independent models across tasks. Experiments on multiple ranking and classification datasets show that the developed method compares to state of the art methods in performance, while at the same time takes significantly less time to train. 

%%%%%%%%%%%%%%%%%%%%%%%%%%%%%%%%%%
%%%%%%%%%%%%%%%%%%%%%%%%%%%%%%%%%%
%%%%%%%%%%%%%%%%%%%%%%%%%%%%%%%%%%
%%%%%%%%%%%%%%%%%%%%%%%%%%%%%%%%%%
%%%%%%%%%%%%%%%%%%%%%%%%%%%%%%%%%%
%%%%%%%%%%%%%%%%%%%%%%%%%%%%%%%%%%
%%%%%%%%%%%%%%%%%%%%%%%%%%%%%%%%%%
%%%%%%%%%%%%%%%%%%%%%%%%%%%%%%%%%%
%%%%%%%%%%%%%%%%%%%%%%%%%%%%%%%%%%
%%%%%%%%%%%%%%%%%%%%%%%%%%%%%%%%%%

\bibliographystyle{plainnat}
\bibliography{gbfs}

\newpage
% !TEX root = gtgbm_aistats.tex
\appendix
\addcontentsline{toc}{section}{Appendices}
\section*{Appendix}

\section{Pseudocode for A-GBM} \label{app:agbm}
Algorithm \ref{agbm} has the pseudocode for the AGBM procedure introduced in Section \ref{sec:setup}. The GBFS training procedure is identical, except with the function being optimized being un-normalized. 

\begin{algorithm}[!h]
  \caption{Pseudocode for A-GBM}
   \label{agbm}
%  %Regularization parameter:\lambda, Step size $\eta$}
  \begin{algorithmic}[1]
  \REQUIRE  Data $\{x_{i},y_{i}\},\,i=1,..,n,$ shrinkage
$\epsilon$, iterations $N,$ penalty parameter $\mu$, tree growth parameter $\alpha$
    \STATE model $H = 0$, residues $g_{i}=y_{i}$, $i=1,2,..,n$.
and selected feature set $\Omega=\emptyset$
  \FOR{$k=1,2,\ldots N$}
 \STATE Fit a tree $h_k$ using $\mu$ to minimize \eqref{gbfs} \label{minfunc} in every split and  $\alpha$ as stopping criteria
 \STATE  $H = H + \epsilon h_k$
 \STATE  $g_{i}=y_{i}-H({x}_{i})$
 \STATE  $\Omega=\Omega\cup\left\{ j,\,\text{ tree }h_{k}\text{uses feature }f_{j}\right\} $
   \ENDFOR 
  \STATE Output  $H$ and $\Omega$
   \end{algorithmic}
\end{algorithm}

\section{Theoretical analysis of GTGBM}
\label{app:theoretical analysis of GTGBM}
\subsection{Notations and Setup}

Consider $(\boldsymbol{X},Y)\sim\mathbb{P}$. $Y$ is the label and
we have $d$ features: $\boldsymbol{X}=(X_{1},..,X_{d})$. $X_{1},,.,X_{d}$
are independent with each other (not assuming have the same distribution)
and $0\leq X_{i}\leq1$ (as GTGBM first standardizes the feature value
to be within $[0,1]$ ). Assume there is an unknown subset $S^{*}\subset[d],$$|S^{*}|=s$,
such that 
\begin{equation}
Y=\mu+\sum_{i\in S^{*}}f_{i}(X_{i})+\epsilon,\label{eq:additive form 1}
\end{equation}
where $\mu=\mathbb{E}Y$ is the population mean and $\epsilon$ is
noise that has mean $0$ and is independent with $\boldsymbol{X}$.
$f_{i}$s are unknown univariate functions. To make the model identifiable,
we can assume without loss of generality that
\begin{equation}
\mathbb{E}f_{i}(X_{i})=0,i\in S^{*}\label{eq: additive form 2}
\end{equation}
This is called a sparse additive model. For
the set-ups of GTGBM, we independently generated $p=\left\lceil es\log(\frac{s}{\delta})\right\rceil $
random subsets of $[d]$: $S_{1},..,S_{p}$, where $e=2.71828..$
is the base of natural logarithm and $\delta\in(0,1)$. From Theorem
\ref{thm:speed}, with high probability ($\geq1-\delta$), for every relevant features
($X_{i}$,$i\in S^{*}$ ), there is a random subset that exactly covers
this feature.

\subsection{Proof of Theorem \ref{thm:speed}}
\label{app:proofspeed}
\begin{proof}
	Suppose we have $d$ features, and without loss of generality  the active features are $f_1, \ldots , f_s \in \{1, 2, \ldots, d \}$. We generate iid subsets $S_1, \ldots, S_p \subset [d]$, such that $\forall j \in [d], \quad P(j \in S_i) = 1/s$.  We want to  show that the probability that exactly one of the relevant features lies in one of the random groups we create is larger than $1 - \delta$. We do this by obtaining an upper bound on it's complement. For convenience, we use the following shorthands: $\{ f_1, \ldots, f_s \} := \Omega, ~\ \{S_1, \ldots, S_p \} := \mathcal{S}$. We bound the probability of the complement of the event we are interested as follows:
	
	\small
	\begin{align}
	\notag
	&P(\exists j \in \Omega ~\ : \forall S \in \mathcal{S}, j \notin S ~\ \textbf{OR} ~\ \exists j' \neq j : j' \in S, ~\ j' \in \Omega) \\
	\notag
	&\leq s (1 - P(f_1 \in S_1 ~\ \textbf{and} ~\ \forall j' \neq f_1, ~\ j' \in \Omega, ~\ j' \notin S_1))^p \\
	\notag
	&= s \left( 1 - \frac{1}{s} \left( 1 - \frac{1}{s} \right)^{s-1} \right)^p \\
	\notag
	& \leq s \exp\left( - \frac{p}{s} \left( 1 - \frac{1}{s} \right)^{s-1}  \right) \\
	\label{finbound}
	& \leq s \exp \left( -\frac{p}{es} \right) \leq \delta
	\end{align}
	
	Where the first inequality follows from the union bound, the second inequality follows from Bernoulli's inequality. The final inequality in  \eqref{finbound} holds so long as $p$ satisfies the condition in the statement of the Theorem. 
\end{proof}

\subsection{Theoretical split criterion in GTGBM}

A key component of tree algorithms are the rules for splitting a node.
For the classical CART algorithm, we greedily build the tree by splitting
with a feature and a threshold such that in the child nodes the sample
are most homogeneous measured by square error loss. Mathematically,
the population version of the split criterion can be written as a
function $L(Z,t)$ of split feature $Z$ (including the ``peusdo''
feature created by GT-GBM) and threshold $t\in\mathbb{R}$:
\begin{eqnarray}
L(Z,t) & = & \mathbb{E}[\left(Y-\mathbb{E}\left[Y|Z<t\right]\right)^{2}1\{Z<t\}\nonumber \\
&&+\left(Y-\mathbb{E}\left[Y|Z\geq t\right]\right)^{2}1\{Z\geq t\}]\label{eq: theoretical split 1}
\end{eqnarray}
Note that the split function is invariant with a shift of a constant
in $Y$, so we may assume $\mu=\mathbb{E}Y=0$ without loss of generality.
Then some calculations lead to 
\begin{equation}
L(Z,t)=\mathbb{E}Y^{2}-\frac{\mathbb{E}^{2}\left[Y1\{Z<t\}\right]}{\mathbb{P}\left(Z<t\right)}-\frac{\mathbb{E}^{2}\left[Y1\{Z\geq t\}\right]}{\mathbb{P}\left(Z\geq t\right)}\label{eq: theoretical split 2}
\end{equation}
Since $\mathbb{E}Y=0$, we have $$\mathbb{E}\left[Y1\{Z<t\}\right]=-\mathbb{E}\left[Y1\{Z\geq t\}\right].$$
Let $M(Z,t)=\mathbb{E}\left[Y1\{Z\geq t\}\right]$, we can further
write 
\begin{equation}
L(Z,t)=\mathbb{E}Y^{2}-\frac{M^{2}(Z,t)}{\mathbb{P}\left(Z<t\right)\mathbb{P}\left(Z\geq t\right)}\label{eq: theoretical split 3}
\end{equation}
In the algorithm, we will choose $(Z,t)$ that minimize $L(Z,t)$
(the sample estimated version, see next section) which is equivalent
to maximize $\frac{M^{2}(Z,t)}{\mathbb{P}\left(Z<t\right)\mathbb{P}\left(Z\geq t\right)}$.
Note that if $Z$ and $Y$ are independent, then 
\begin{align}
M(Z,t) & =\mathbb{E}\left[Y1\{Z\geq t\}\right]\nonumber\\
& =\mathbb{E}\left[Y\right]\mathbb{P}\left(Z\geq t\right)\nonumber\\
& =0 \label{eq: theoretical split 3}
\end{align}
Thus no variance reduction takes into place. Let's recall the GTGBM
procedure to find the split feature: for the $p$ independently generated
random group of features, we perform binary search. That is, for random
subset $S\subset[d],$write 
\[
Z_{S}=\sum_{i\in S}X_{i},
\]
we split $S$ into left-half $S_{L}$ and right-half $S_{R}$ and
calculate $\inf_{t}L(Z_{S_{L}},t)$ and $\inf_{t}L(Z_{S_{R}},t)$
. We select the half with smaller value and recursively find the candidate
split feature. We find the candidate split features for all $p$ random
subsets of features, and we choose the best split feature among them.
Now we show that, if we have access to the theoretical split criterion
(that corresponds to the ideal situation that we have infinite amount
of data), the GTGBM split-finding procedure can actually find the
best split feature. We only need to show that all relevant features:
$X_{i}$,$i\in S^{*}$ are among the candidate split features. For
$i\in S^{*}$, from Theorem \ref{thm:speed} we know that there is a random subset
$S\in\{S_{1,..,}S_{p}\}$ such that $i\in S$ and for any $i'\in S^{*},i'\neq i$,
we have $i'\notin S$. Now we show that when we perform binary search
on $S$, the half that contains the important feature index $i$ is
always been selected. Thus the output of binary search on $S$ is
exactly this index $i$. Suppose the left half $S_{L}$ contains $i$.
Then $S_{R}$ doesn't contain $i$ and also doesn't contain any $i'\in S^{*},i'\neq i$
since $S$ doesn't contain them. Thus $Z_{S_{R}}$ is independent
with $Y$, so $M(Z_{S_{R}},t)=0$ for any $t$. On the other hand
\begin{eqnarray}
M(Z_{S_{L}},t) & = & \mathbb{E}\left[Y1\{Z_{S_{L}}\geq t\}\right]\nonumber \\
& = & \sum_{i\in S^{*}}\mathbb{E}\left[f_{i}(X_{i})1\{Z_{S_{L}}\geq t\}\right]\nonumber \\
& = & \mathbb{E}\left[f_{i}(X_{i})1\{X_{i}+\sum_{i'\neq i,i'\in S_{L}}X_{i'}\geq t\}\right]\label{eq: the half contain important feature}
\end{eqnarray}
We can choose $t$ such that $M(Z_{S_L},t)\neq0$, as long as $f_{i}$
is not degenerated. Thus we always have 
\begin{eqnarray*}
	\inf_{t}L(Z_{S_{L}},t) & \leq & L(Z_{S_{L}},t)\\
	& = & \mathbb{E}Y^{2}-\frac{M^{2}(Z_{S_{L}},t)}{\mathbb{P}\left(Z_{S_{L}}<t\right)\mathbb{P}\left(Z_{S_{L}}\geq t\right)}\\
	& < & \mathbb{E}Y^{2}\\
	& = & \inf_{t}L(Z_{S_{R}},t).
\end{eqnarray*}
But in reality, we are using sample version of split function that
only approximates the theoretical split function. So the condition
for GTGBM to successfully find the best split feature depends on how
the approximation error between theoretical split function and empirical
split function and magnitude of $M^{2}(Z_{S_L},t)$ (still assumes $S_{L}$
is the half that contains the relevant feature index) change with
sample size $n$ at a node and total number of features $d$. Intuitively,
the increase of dimension $d$ will harm the signal strength $M^{2}(Z_{S_L},t)$
since the irrelevant part $\sum_{i'\neq i,i'\in S_{L}}X_{i'}$ in
equation (\ref{eq: the half contain important feature})becomes more
dominant. We rigorously showed that (see lemma \ref{app: therotical split lemma}), under fairly general
condition we have 
\begin{equation}
M^{2}(Z_{S_{L}},t)\gtrsim\frac{1}{\left|S_{L}\right|}\gtrsim\frac{s}{d}.\label{eq: signal decay 1}
\end{equation}
Then we just need to know how well we can approximate theoretical
split function by the empirical ones with sample size $n$.

\subsection{Empirical split criterion in GTGBM}

Suppose we have $i.i.d$ sample in a node $(\boldsymbol{X}_{i},Y_{i})\sim\mathbb{P},i=1,2,..,n$.
$\boldsymbol{X}_{i}=(X_{i1},..,X_{id})$. The empirical split function
is 
\begin{equation}
L_{n}(Z,t)=\frac{1}{n}\left(\sum_{i:Z_{i}<t}\left(Y_{i}-\bar{Y}_{L}\right)^{2}+\sum_{i:Z_{i}\geq t}\left(Y_{i}-\bar{Y}_{R}\right)^{2}\right)\label{eq: empirical split 1}
\end{equation}
where $\bar{Y}_{L}=\frac{\sum_{i}Y_{i}1\{Z_{i}<t\}}{\sum_{i}1\{Z_{i}<t\}}$
, $\bar{Y}_{R}=\frac{\sum_{i}Y_{i}1\{Z_{i}\geq t\}}{\sum_{i}1\{Z_{i}\geq t\}}$
and $Z_{i}$, $i=1,..,n$ is the $i.i.d$ sample for split feature
$Z$. With a standard argument and concentration inequality
(see lemma \ref{app: emprical split lemma} ), we
can prove 
\begin{equation}
\sup_{t}\left|L_{n}(Z,t)-L(Z,t)\right|=O_p(\frac{1}{\sqrt{n}}).\label{eq: emprical approx 1}
\end{equation} Thus with high probability,
we have 
\begin{eqnarray}
\inf_{t}L_{n}(Z_{S_{L}},t) & \leq & \inf_{t}L(Z_{S_{L}},t)+O(\frac{1}{\sqrt{n}})\nonumber \\
& \lesssim & \mathbb{E}Y^{2}-\frac{s}{d}+O(\frac{1}{\sqrt{n}})\nonumber \\
& = & \inf_{t}L(Z_{S_{R}},t)-\frac{s}{d}+O(\frac{1}{\sqrt{n}})\nonumber \\
& \leq & \inf_{t}L_{n}(Z_{S_{R}},t)-\frac{s}{d}+O(\frac{1}{\sqrt{n}})\label{eq: emprical split fun compare 1}
\end{eqnarray}
The first and last inequality is from (\ref{eq: emprical approx 1})
and the second inequality is from (\ref{eq: signal decay 1}).  So, we only need $n\gtrsim (\frac{d}{s})^2$ for GTGBM to find the best split variables. 
\begin{comment}
Thuswe can choose $x=\frac{s}{4d}$ so that even empirically we are able
to select the half of group that contain the relevant feature index
with high probability. This binary comparison happens at most $\log_{2}|S|\lesssim\log(\frac{d}{s})$
for each of $m=\left\lceil es\log(\frac{s}{\delta})\right\rceil $
random subsets, so by a union bound, the overall failure probability
is upper bounded by 
\begin{eqnarray*}
	\delta+s\log(\frac{s}{\delta})\log(\frac{d}{s})O(\frac{\exp(-nx^{2})}{x}) & = & \delta+O\left(s\log(\frac{s}{\delta})\log(\frac{d}{s})\exp\left(-n(\frac{s}{d})^{2}-\log(\frac{s}{d})\right)\right)
\end{eqnarray*}
If we assume $n\gtrsim(\frac{d}{s})^{2}\log(\frac{d}{s})$, and decay
$\delta$ to zero with the increase of $n$, say $\delta=O(\frac{1}{n})$,
we have 
\begin{eqnarray*}
	\delta+O\left(s\log(\frac{s}{\delta})\log(\frac{d}{s})\exp\left(-n(\frac{s}{d})^{2}-\log(\frac{s}{d})\right)\right) & \lesssim & \delta+O\left(\frac{s\log(\frac{s}{\delta})\log(\frac{d}{s})}{n}\right)\\
	& \lesssim & O\left(\frac{s\log(s)\log(\frac{d}{s})+\log n}{n}\right)
\end{eqnarray*}
which goes to zero as $n$ tends to infinity.
\end{comment} So, we 
\subsection{Proof of Theorem \ref{thm:correct}}
\label{app:proofcorrect}
The above subsections did some intuitive calculations that motivate the
claim. This subsection aims at providing rigorous statement and filling
the gaps.
First let's recall the conditions assumed in theorem \ref{thm:correct}.

Assume
\begin{enumerate}
\item \emph{$X_{i}$ has bounded probability density function $p_{i}(x)$ and positive variance.
Denote $B_{d}^{2}=\text{Var}(X_{1}+..+X_{d})$. Suppose $B_{d}\rightarrow\infty,d\rightarrow\infty$}.
%and the limit 
%\[
%\eta=\lim_{d\rightarrow\infty}\frac{\mu_{d}}{B_{d}}
%\]
%exists, where $\mu_{d}=\mathbb{E}\left[X_{1}+..+X_{d}\right]$.}
\item \emph{The unknown functions in (\ref{eq:additive form 1}) are bounded
monotone functions.}
\end{enumerate}

We have following two lemmas:

\begin{lemma}
\label{app: therotical split lemma}
Recall the notation, for subset $S\subset[d]$,
$Z_{S}=\sum_{i\in S}X_{i}$. if there is an index $i\in S^{*}$ that
$i\in S$ and for any $i'\neq i,i'\in S^{*}$ we have $i'\notin S$. Denote $S'=S\backslash\{i\}$ .
Assume the unknown function component $f_{i}$ is bounded monotone.
Also assume condition 1 in theorem 1. Then there exists constants
$t_{0}$ , $d_{0}>0,c_{0}>0$ that only depend on the unknown functions
in (\ref{eq:additive form 1}) %and $\eta,$%
such that when $|S|\geq d_{0}$,
we have 
\begin{equation}
L(Z_{S},t_{0}+\mathbb{E}Z_{S'})\leq\mathbb{E}Y^{2}-\frac{c_{0}}{|S|}\label{eq: density lemma 1}
\end{equation}
\end{lemma}
\begin{proof}[proof of lemma \ref{app: therotical split lemma}]
	From (\ref{eq: theoretical split 3}), we only need to show
	that\emph{ }there exists constants $t_{0}$ , $d_{0}>0,c_{0}>0$ ,
	such that 
	\begin{equation}
	\frac{M^{2}(Z_{S},t_{0}+\mathbb{E}Z_{S'})}{\mathbb{P}\left(Z_{S}<t_{0}+\mathbb{E}Z_{S'}\right)\mathbb{P}\left(Z_{S}\geq t_{0}+\mathbb{E}Z_{S'}\right)}\geq\frac{c_{0}}{|S|}\label{eq: density lemma 2}
	\end{equation}
	First let's look at the numerator. From (\ref{eq: the half contain important feature}),
	we have 
	\begin{equation}
	M(Z_{S},t)=\mathbb{E}\left[f_{i}(X_{i})1\{X_{i}+\sum_{i'\neq i,i'\in S}X_{i'}\geq t\}\right]\label{eq:density lemma 3}
	\end{equation}
	Denote %$S'=S\backslash\{i\}$ and%
	$X_{i}$ and $Z_{S'}-\mathbb{E}Z_{S'}$ 's probability
	density function as $p_{i}(x)$ and $\bar{p}_{Z_{S'}}(z)$ respectively.
	Since $X_{i}$ and $Z_{S'}$ are independent, we have 
	\begin{align}
	&\mathbb{E}\left[f_{i}(X_{i})1\{X_{i}+Z_{S'}-\mathbb{E}Z_{S'} \geq t\}\right]  \nonumber \\
	& =  \int_{x+z\geq t}f_{i}(x)p_{i}(x)\bar{p}_{Z_{S'}}(z)dzdx \nonumber \\
	& =  \int f_{i}(x)p_{i}(x)\int_{z\geq t-x}\bar{p}_{Z_{S'}}(z)dzdx \nonumber
	\end{align}
	On the other hand, since $f_{i}$ is monotone function (without loss
	of generality assume it's monotone increasing), then there exists
	$t_{0}\in[0,1]$ such that $f_{i}(t_{0})=0$ and $f_{i}(t)>0$ for
	$t>t_{0}$ and $f_{i}(t)<0$ for $t<t_{0}$. Then, $\mathbb{E}\left[f_{i}(X_{i})1\{X_{i}+Z_{S'}-\mathbb{E}Z_{S'}\geq t_{0}\}\right]$
	can be written as 
	\begin{multline}
	\int_{x\geq t_{0}}f_{i}(x)p_{i}(x)\int_{z\geq t_{0}-x}\bar{p}_{Z_{S'}}(z)dzdx\\ 
	+\int_{x<t_{0}}f_{i}(x)p_{i}(x)\int_{z\geq t_{0}-x}\bar{p}_{Z_{S'}}(z)dzdx\\
	=\int_{1\geq x\geq t_{0}}f_{i}(x)p_{i}(x)\int_{t_{0}-x}^{0}\bar{p}_{Z_{S'}}(z)dzdx\\
	-\int_{0\leq x<t_{0}}f_{i}(x)p_{i}(x)\int_{0}^{t_{0}-x}\bar{p}_{Z_{S'}}(z)dzdx\label{eq: density lemma 5}
	\end{multline}
	The equation is from the fact that $\int_{x\geq t_{0}}f_{i}(x)p_{i}(x)dx+\int_{x<t_{0}}f_{i}(x)p_{i}(x)dx=\mathbb{E}\left[f_{i}(X_{i})\right]=0.$  Let $m_{Z_{S'}}=\min_{z\in[t_{0}-1,t_{0}]}\bar{p}_{Z_{S'}}(z)$. Then the
	right hand side of (\ref{eq: density lemma 5}) is lower bounded by
	\begin{equation}
	m_{Z_{S'}}\intop_{0}^{1}(x-t_{0})f_{i}(x)p_{i}(x)dx.\label{eq: density lemma 6-1}
	\end{equation}
	Note that $(x-t_{0})f_{i}(x)p_{i}(x)\geq0$ for any $x\in[0,1]$ and
	there exists a positive measure set such that $(x-t_{0})f_{i}(x)p_{i}(x)>0$
	(otherwise $X_{i}$ is degenerated). Thus we denote $v_{0}=\intop_{0}^{1}(x-t_{0})f_{i}(x)p_{i}(x)dx$
	and $v_{0}>0$. Now let's look at the other factor $m_{Z_{S'}}$ in
	(\ref{eq: density lemma 6-1}) Denote $\tilde{Z}_{S'}=\frac{Z_{S'}-\mathbb{E}Z_{S'}}{\sqrt{\text{Var}(Z_{S'})}}$
	as standardized $Z_{S'}$ , then we have 
	\begin{equation}
	\bar{p}_{Z_{S'}}(z)=\frac{1}{\sqrt{\text{Var}(Z_{S'})}}p_{\tilde{Z}_{S'}}(\frac{z}{\sqrt{\text{Var}(Z_{S'})}}).\label{eq: density function scale-1}
	\end{equation}
	From condition 1 and the well known local limit theorem, the standardized density function $p_{\tilde{Z}_{S'}}(z)$
	uniformly converge to standardized normal density $\phi(z)$ as $|S'|\rightarrow\infty$.
	Moreover 
	\begin{equation}
	\lim_{|S'|\rightarrow\infty}\sqrt{\text{Var}(Z_{S'})}m_{Z_{S'}}=\phi(0)\label{eq: density lemma 7-1}
	\end{equation}
	since from condition 1, we have $\lim_{|S'|\rightarrow\infty}\frac{z}{\sqrt{\text{Var}(Z_{S'})}}=0, \forall z \in [t_0-1,t_0]$. 
	Combined with (\ref{eq:density lemma 3})(\ref{eq: density lemma 5})(\ref{eq: density lemma 6-1}),
	we conclude that there exists a constant $d_{1}$ such that when $|S|>d_{1}$,
	we have 
	\begin{equation}
	M^{2}(Z_{S},t_{0}+\mathbb{E}Z_{S'})\geq\frac{v_{0}^{2}\phi^{2}(0)}{2\text{Var}(Z_{S'})}\geq\frac{v_{0}^{2}\phi^{2}(0)}{2|S|}\label{eq: density lemma 8}
	\end{equation}
	where the second inequality follows from $\text{Var}(Z_{S'})=\sum_{i\in S'}\text{Var}(X_{i})\leq|S'|<|S|$
	since $X_{i}\leq1$. For the denominator in (\ref{eq: density lemma 2}),
	from Central Limit Theorem, we have 
	\begin{align*}
	\mathbb{P}\left(Z_{S}<t_{0}+\mathbb{E}Z_{S'}\right) & =\mathbb{P}\left(\frac{Z_{S}-\mathbb{E}Z_{S}}{\sqrt{\text{Var}(Z_{S})}}<\frac{t_{0}-\mathbb{E}X_{i}}{\sqrt{\text{Var}(Z_{S})}}\right)\rightarrow\Phi(0)=\frac{1}{2}
	\end{align*}
	as $|S|\rightarrow\infty$,
	where $\Phi$ is the distribution function of standard normal. Thus
	there exists a constant $d_{2}$, such that when $|S|>d_{2}$, we
	have 
	\begin{equation}
	\mathbb{P}\left(Z_{S}<t_{0}\right)\mathbb{P}\left(Z_{S}\geq t_{0}\right)\leq2\Phi^{2}(0)=\frac{1}{2}.\label{eq: density lemma 9}
	\end{equation}
	Thus combine (\ref{eq: density lemma 8})(\ref{eq: density lemma 9}),
	we showed that for $|S|>d_{0}=\max\{d_{1},d_{2}\}$, we have 
	\[
	\frac{M^{2}(Z_{S},t_{0}+\mathbb{E}Z_{S'})}{\mathbb{P}\left(Z_{S}<t_{0}+\mathbb{E}Z_{S'}\right)\mathbb{P}\left(Z_{S}\geq t_{0}+\mathbb{E}Z_{S'}\right)}\geq\frac{c_{0}}{|S|}
	\]
	where $c_{0}=v_{0}^{2}\phi^2(0)>0$ . That
	concludes the proof.
\end{proof}

\begin{lemma}
	\label{app: emprical split lemma}
	There exists positive constants $c_1,c_2$ that only depend on the unknown fixed component functions such that for any $0<x<1$
	\begin{equation}
	\mathbb{P}\left(\sup_{t}\left|L_{n}(Z,t)-L(Z,t)\right|\leq x\right)\geq1-c_{1}\exp(-c_{2}nx^{2})\label{eq: deviation lemma 1}
	\end{equation}
\end{lemma}

\begin{proof}[proof of lemma \ref{app: emprical split lemma}]
Let $\mu_L=\frac{\mathbb{E}[Y1{Z<t}]}{\mathbb{P}(Z<t)}$, $\mu_R=\frac{\mathbb{E}[Y1{Z\geq t}]}{\mathbb{P}(Z\geq t)}$ and $n_L=\sum_{i}1\{Z_i<t\}$, $n_R=\sum_{i}1\{Z_i\geq t\}$ .  Define 
\begin{equation}
\tilde{L}_{n}(Z,t)=\frac{1}{n}\left(\sum_{i:Z_{i}<t}\left(Y_{i}-\mu_L\right)^{2}+\sum_{i:Z_{i}\geq t}\left(Y_{i}-\mu_R\right)^{2}\right)\label{eq: deviation lemma 2}
\end{equation}
Then 
\begin{multline}
\tilde{L}_{n}(Z,t)-L_{n}(Z,t)=\\ 
\frac{1}{n}\sum_{i:Z_{i}<t}(\bar{Y}_L-\mu_L)(2Y_i-\mu_L-\bar{Y}_L)\\
	+\frac{1}{n}\sum_{i:Z_{i}\geq t}(\bar{Y}_R-\mu_R)(2Y_i-\mu_R-\bar{Y}_R)\\
	=\frac{n_L}{n}(\bar{Y}_L-\mu_L)^2+\frac{n_R}{n}(\bar{Y}_R-\mu_R)^2\label{eq: deviation lemma 3}
\end{multline}
Also we can write $\bar{Y}_L-\mu_L$ as 
\begin{multline}
\frac{1}{n_L}\sum_{i}(Y_i1\{Z_i<t\}-\frac{\mathbb{E}[Y1\{Z<t\}]}{\mathbb{P}(Z<t)})\\
=\frac{n}{n_L}\frac{1}{n}\sum_{i}(Y_i1\{Z_i<t\}-\mathbb{E}[Y1\{Z<t\}])\\
+\mathbb{E}[Y1\{Z<t\}](\frac{n}{n_L}-\frac{1}{\mathbb{P}(Z<t)})\label{eq: deviation lemma 4}
\end{multline}
Since $1\{Z_i<t\}-\mathbb{P}(Z<t)$ and $Y_i1\{Z_i<t\}-\mathbb{E}[Y1\{Z<t\}]$ are i.i.d mean 0 bounded random variables (and the bound doesn't depend on $t$), from Bernstain inequality, for any $t$ and $x>0$, we have 
\begin{equation}
\mathbb{P}\left(\frac{1}{n}\left|\sum_{i}1\{Z_i<t\}-\mathbb{P}(Z<t)\right|\geq x\right)\leq 2\exp(-c_{1}nx^{2})\label{eq: deviation lemma 5}
\end{equation}
and
\small
\begin{equation}
\mathbb{P}\left(\frac{1}{n}\left|\sum_{i}Y_i1\{Z_i<t\}-\mathbb{E}[Y1\{Z<t\}]\right|\geq x\right)\leq 2\exp(-c_{2}nx^{2})\label{eq: deviation lemma 6}
\end{equation}
where $c_1,c_2$ are positive constants that don't depend on $t$.  Combine (\ref{eq: deviation lemma 4})(\ref{eq: deviation lemma 5})(\ref{eq: deviation lemma 6}), with proper change of the constants $c_1,c_2$, we conclude that, for all $t$ and any $x>0$
\begin{equation}
\mathbb{P}\left(\left|\bar{Y}_L-\mu_L\right|\geq x\right)\leq c_1\exp(-c_{2}nx^{2})\label{eq: deviation lemma 7}
\end{equation}
We can apply the same argument to $\bar{Y}_R-\mu_R$. Thus for any $x>0$,
\begin{equation}
\mathbb{P}\left(\sup_{t}\left|\tilde{L}_{n}(Z,t)-L_{n}(Z,t)\right|\geq x\right)\leq c_1\exp(-c_{2}nx)\label{eq: deviation lemma 8} 
\end{equation}
for proper constants $c_1,c_2$. When $x<1$, the right hand side of (\ref{eq: deviation lemma 8}) $\leq c_1\exp(-c_{2}nx^2)$.
Thus we only need to prove 
\begin{equation}
\mathbb{P}\left(\sup_{t}\left|\tilde{L}_{n}(Z,t)-L(Z,t)\right|\geq x\right)\leq c_1\exp(-c_{2}nx^2)\label{eq: deviation lemma 9} 
\end{equation}
This also follows from Bernstain inequality, since $ \tilde{L}_{n}(Z,t)-L(Z,t)$ is the average of i.i.d mean 0 random variables $$w_i:=\left(Y_{i}-\mu_L\right)^{2}1\{Z_i<t\}+\left(Y_{i}-\mu_R\right)^{2}1\{Z_i\geq t\}-L(Z,t)$$ $w_i$ is also bounded (since $Y_i$ are bounded) and the bound doesn't depend on $t$. 
\end{proof}

Now let's go back to the proof of main theorem.
From (\ref{eq: emprical split fun compare 1}) and lemma \ref{app: therotical split lemma} and \ref{app: emprical split lemma}, the failure probability of identifying the correct half group that contains the important feature is bounded by $c_1\exp(-c_2nx^2)$  with $x=\frac{c_0s}{4d}$. Given $\delta\in(0,1)$, since GTGBM performs at most $es\log(\frac{2s}{\delta})\log_{2}(\frac{d}{s})$ times of comparing two splitted groups of variables (assume we generate $es\log(\frac{2s}{\delta})$ random subsets) , by union bound and theorem \ref{thm:speed}, the overal failure probability is bounded by $$\frac{\delta}{2}+c_1es\log(\frac{2s}{\delta})\log_{2}(\frac{d}{s})\exp(-c_2nx^2)$$  with $x=\frac{c_0s}{4d}$. Solve $n$ for $$c_1es\log(\frac{2s}{\delta})\log_{2}(\frac{d}{s})\exp(-c_2nx^2)\leq \frac{\delta}{2}$$ with $x=\frac{c_0s}{4d}$ gives the conclusion.

\section{Optimal Hyperparameters to Reproduce Results on Public Datasets}
\label{app:exptresults}
Here we give additional details required to reproduce the results we obtained on all 3 public datasets. We used the train/test split that was provided online in all the cases: $6000/1000$ for Gisette, $80000,20000$ for Epsilon and $100K, 100K$ for Flight Delay

For tuning the hyperparameters, we further split the train set into an 80-20 train and validation set, and cross-validate on the latter. Table \ref{tab:params} lists the optimal hyperparameters for all the algorithms used. `$\alpha$' is the minimum fraction of data in an internal node (parameter that controls the size of a single tree). %For LightGBM, the corresponding parameter is the actual number of samples, so the value in the table is after computing the corresponding $\alpha$
\small
\begin{table}[!h]
\caption{Optimal hyperparameters for all methods  \label{tab:params}}
\begin{tabular}{|c|c|c|c|c|}
\hline
\textbf{Dataset}	& \textbf{Method}	& \textbf{$\mu$}	& \textbf{shrinkage $\epsilon$}	& \textbf{$\alpha$} \\
\hline
Gisette		&	GBDT		&	-			&	0.1			& 0.02	\\
			& GBFS		&	1.1			&	0.1			& 0.02\\
			& A-GBM		&		0.01		&	0.1			& 0.02\\
			& GT-GBM	&	0.001			&	0.1			& 0.02\\
\hline
Epsilon		&	GBDT	&	-			&	0.1			& 0.02\\
			& GBFS		&	2.0			&	0.1			& 0.02\\
			& A-GBM		&	0.0004			&	0.1			& 0.02\\
			& GT-GBM	&	0.0001			&	0.1			& 0.02\\
\hline
Flight		&	GBDT	&		-		&	0.1			& 0.1\\
			& GBFS		&	4			&	0.1			& 0.1\\ 
			& A-GBM		&	0.0004			&	0.1			& 0.1\\
			& GT-GBM	&	0.0002			&	0.1			& 0.1\\
\hline
\end{tabular}
\end{table}

\section{Performance When All Features Are Used}
\label{app:allfeatures}
For the sake of completeness, we provide the optimum hyperparameter values as well as the results obtained on the public datasets when we use all the available features to train the model.  Note that we report this performance for the sake of comparison, and as we mentioned earlier, such a method is not practical in the applications we consider. The results are provided in Table \ref{app:allfeatures}
\begin{table}[!h]
\caption{Performance of the full GBDT model on all public datasets}
\begin{tabular}{|c|c|c|c|c|}
\hline
\textbf{Dataset}	& \textbf{Method}		& \textbf{shrinkage $\epsilon$}	& \textbf{$\alpha$} & \textbf{AUC}\\
\hline
Gisette		&GBDT-Full			&	0.1			& 0.02		& 99.33\\
\hline
Epslion		&	GBDT-Full			&	0.1			& 0.02 & 92.34 \\
\hline
Flight		&	GBDT-Full			&	0.1			& 0.1 & 71.74\\
\hline
\end{tabular}
\end{table}

\section{Performance on Internal Classification Datasets} \label{app:internal_clf}

For the internal classification dataset, we compute the area under the ROC curve, and the Precision at 2. The task in both cases is to identify items in response to query-item pairs that have been marked as ``incorrect." We see from Table \ref{tab:c} that GBDT-topK methods are suboptimal, and GT-GBM matches or outperforms GBFS, while being vastly superior in terms of training time.

\begin{table}[!h]
\tiny
\caption{\label{tab:c}Comparison of various methods for the classification tasks (C1 and C2). In both cases, GBDT-topK is suboptimal, and GT-GBM narrowly outperforms GBFS. Bold numbers indicate the best result.}

\begin{tabular}{|c|c|c|c|c|}
\hline 
 \textbf{Dataset} & \textbf{Measure}  & \textbf{GBDT-topK}  & \textbf{GBFS}  & \textbf{GT-GBM} \tabularnewline
\hline 
\hline 
\textbf{C1} & \textbf{AUC\_ROC}  & 0.918  & \textbf{0.922}  & 0.920  \tabularnewline
\hline 
 & \textbf{prec@k=2}  & 0.751  & 0.770  & \textbf{0.773}  \tabularnewline
\hline 
 & \textbf{RMSE}  & 0.260 & \textbf{0.258}  & \textbf{0.258} \tabularnewline
\hline
\hline 
\textbf{C2} & \textbf{AUC\_ROC}  & 0.910  & 0.910  & \textbf{0.912} \tabularnewline
\hline 
&\textbf{prec@k=2}  & 0.874  & 0.875  & \textbf{0.878} \tabularnewline
\hline 
&\textbf{RMSE}  & 0.219  & \textbf{0.218}  & \textbf{0.218} \tabularnewline
\hline 
\end{tabular}
\end{table}

\section{Multitask Results on M2}
\label{app:m2}
\begin{figure}
\centering
\includegraphics[width = 70mm, height = 40mm]{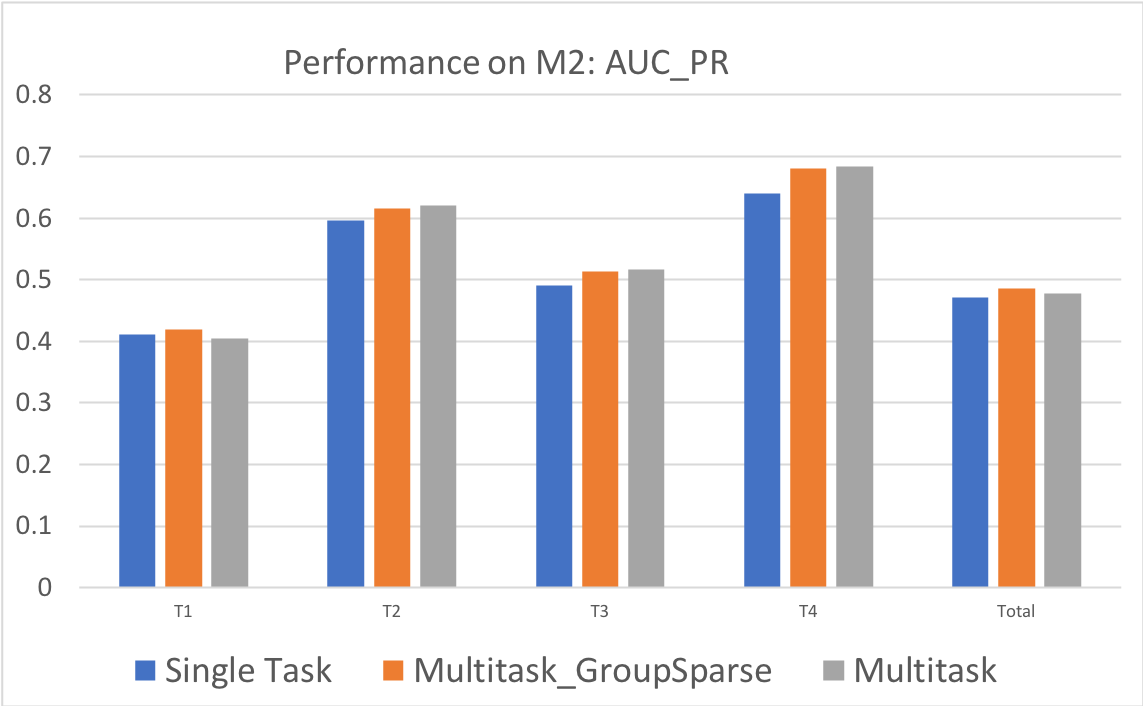} 
\caption{Performance on M2, for Area Under Precision-Recall curves. As in the previous experiment, using both task-specific and across-task features is beneficial. The performance boosts for tasks T2-T4 arise from using the data from T1, which has the largest and cleanest dataset. }
\label{mtlm2}
\end{figure}

There are 4 countries in total. Again, we hypothesize that there will be features that might be common across countries that we can use, and country specific features that depend on the items available, and vagaries of the languages spoken in those countries. We aim to see if combining information from various sources and training joint models helps to achieve better metrics as compared to training models individually. Figure \ref{mtlm2} again shows that the multitask GTGBM outperforms the single task and traditional multitask counterparts. 

\end{document}